\documentclass[runningheads]{llncs}
\usepackage{graphicx}

\usepackage{multirow}
\usepackage{graphicx}
\usepackage{subfigure}
\usepackage{ulem}
\usepackage{url}
\usepackage{comment}
\usepackage{marvosym}

\begin{document}

\title{AutoLaparo: A New Dataset of Integrated Multi-tasks for Image-guided Surgical Automation in Laparoscopic Hysterectomy}
% \author{Paper ID: 1988}
\titlerunning{AutoLaparo: Dataset of Integrated Multi-tasks for Hysterectomy}
\authorrunning{Z. Wang et al.}

\author{Ziyi Wang\inst{1,2} \and
Bo Lu\inst{1,2,3}\and
Yonghao Long\inst{4}\and 
Fangxun Zhong\inst{1,2}\and \\
Tak-Hong Cheung\inst{5}\and
Qi Dou\inst{4,2}\and
Yunhui Liu\inst{1,2}\textsuperscript{(\Letter)}}

% index{Wang, Ziyi}
% index{Lu, Bo}
% index{Long, Yonghao}
% index{Zhong, Fangxun}
% index{Cheung, Tak-Hong}
% index{Dou, Qi}
% index{Liu, Yunhui}


\institute{Department of Mechanical and Automation Engineering, \\The Chinese University of Hong Kong\and
T Stone Robotics Institute, The Chinese University of Hong Kong \\  
 \email{yhliu@mae.cuhk.edu.hk}\\ \and
Robotics and Microsystems Center, School of Mechanical and Electric Engineering, \\Soochow University \\ \and
Department of Computer Science and Engineering, \\The Chinese University of Hong Kong\\\and
Department of Obstetrics and Gynaecology, Prince of Wales Hospital, \\The Chinese University of Hong Kong}

\maketitle

\begin{abstract}
Computer-assisted minimally invasive surgery has great potential in benefiting modern operating theatres.
The video data streamed from the endoscope provides rich information to support context-awareness for next-generation intelligent surgical systems.
To achieve accurate perception and automatic manipulation during the procedure, learning based technique is a promising way, which enables advanced image analysis and scene understanding in recent years.
However, learning such models highly relies on large-scale, high-quality, and multi-task labelled data. This is currently a bottleneck for the topic, as available public dataset is still extremely limited in the field of CAI.
In this paper, we present and release the first integrated dataset (named AutoLaparo) with multiple image-based perception tasks to facilitate learning-based automation in hysterectomy surgery. 
Our AutoLaparo dataset is developed based on full-length videos of entire hysterectomy procedures. 
Specifically, three different yet highly correlated tasks are formulated in the dataset, including surgical workflow recognition, laparoscope motion prediction, and instrument and key anatomy segmentation. 
In addition, we provide experimental results with state-of-the-art models as reference benchmarks for further model developments and evaluations on this dataset. 
The dataset is available at \url{https://autolaparo.github.io}.

\end{abstract}

\section{Introduction}

New technologies in robotics and computer-assisted intervention (CAI) are widely developed for surgeries, to release the burden of surgeons on manipulating instruments, identifying anatomical structures, and operating in confined spaces~\cite{barbash2010new,maier2017surgical,taylor2008medical}. 
The advancements of these technologies will thereby facilitate the development of semi- and fully-automatic robotic systems which can understand surgical situations and even make proper decisions in certain tasks.
Nowadays, surgeries conducted in a minimally invasive way, e.g. laparoscopy, have become popular~\cite{tsui2013minimally}, and videos recorded through the laparoscope are valuable for image-based studies~\cite{maier2022surgical,topol2019high}.
To enhance surgical scene understanding for image-guided automation, one promising solution is to rely on learning-based methods.

As deep learning methods are data-driven, a large amount of surgical data is required to train and obtain reliable models for task executions.
Especially in surgical field, due to various types of surgical procedures, corresponding image and video data vary in surgical scenes, surgical workflows, and the instruments used \cite{grammatikopoulou2021cadis,maier2022surgical}.
For laparoscopic minimally invasive surgery (MIS), some datasets have been established and released to improve the learning-based algorithm for different surgical tasks, such as action recognition \cite{gao2014jhu}, workflow recognition \cite{nakawala2019deep}, instrument detection and segmentation \cite{sarikaya2017detection}.
However, these datasets are not collected from real-surgery nor large enough to develop applicable and robust models in practice.
Recently, several research teams have worked on developing dataset at large scales \cite{allan20202018,allan20192017,twinanda2016endonet}, but most are only designed and annotated for one certain task. 
In terms of clinical applicability, data from different modalities are needed to better understand the whole scenario, make proper decisions, as well as enrich perception with multi-task learning strategy
\cite{dergachyova2016automatic,huaulme2022peg}.
Besides, there are few datasets designed for automation tasks in surgical application, among which the automatic laparoscopic field-of-view (FoV) control is a popular topic as it can liberate the assistant from such tedious manipulations with the help from surgical robots \cite{bihlmaier2014automated}.
Therefore, surgical tasks with multiple modalities of data should be formulated and proposed for image-guided surgical automation.

For application, we target laparoscopic hysterectomy, a gynaecologic surgery commonly performed for patients diagnosed with adenomyosis, uterine fibroids or cancer \cite{farquhar2002hysterectomy,merrill2008hysterectomy}.
By now, some obstacles remain in the development of learning-based approaches for surgical automation in laparoscopic hysterectomy as public datasets for this procedure are only sparsely available.
Although several datasets have been presented for endometriosis diagnosis \cite{leibetseder2020glenda}, action recognition \cite{leibetseder2018lapgyn4}, and anatomical structures and tool segmentation \cite{zadeh2020surgai}, they are small in scale of cases or annotations and insufficient for learning-based model development.

In this paper, we present AutoLaparo, the first large-scale dataset with integrated multi-tasks towards image-guided surgical automation in laparoscopic hysterectomy.
Three tasks along with the corresponding data and annotations are designed: Task 1 surgical workflow recognition, where 21 videos of complete procedures of laparoscopic hysterectomy with a total duration of 1388 minutes are collected and annotated into 7 phases, proposed as the first dataset dedicating workflow analysis in this surgery; Task 2 laparoscope motion prediction, where 300 clips are selected and annotated with 7 types of typical laparoscope motion towards image-guided automatic FoV control; and Task 3 instrument and key anatomy segmentation, where 1800 frames are sampled from the above clips and annotated with 5936 pixel-wise masks of 9 types, providing rich information for scene understanding.
It is worth noting that these tasks, data and annotations are highly correlated to support multi-task and multi-modality learning for advanced surgical perception towards vision-based automation.
In addition, benchmarks of state-of-the-art methods for each task are presented as reference to facilitate further model development and evaluation on this dataset.

\begin{figure*}[t]
\centering
\includegraphics[width=\textwidth]{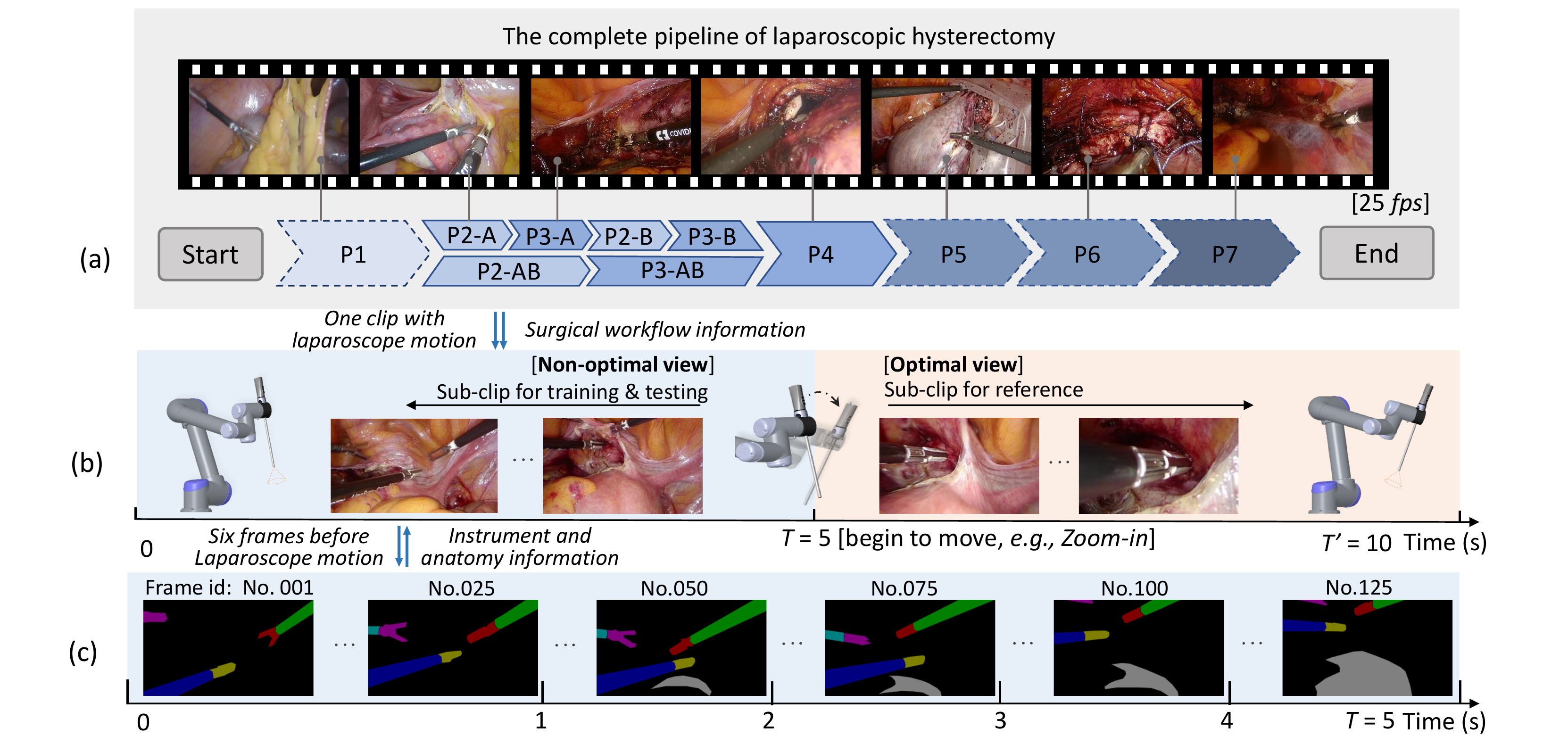}
\caption{Overview of AutoLaparo, the proposed integrated dataset of laparoscopic hysterectomy.
(a) Order information of the procedure with sample frames of each phase (P1-P7). 
(b) One clip that contains the "\textit{Zoom-in}" motion of laparoscope. 
(c) Six frames are sampled from the above clip and pixel-wise annotated for segmentation.}
\label{flow_order}
\end{figure*}

\section{Dataset Design and Multi-tasks Formulation}
Towards image-guided surgical automation, we propose AutoLaparo, integrating multi-tasks to facilitate advanced surgical scene understanding in laparoscopic hysterectomy, as illustrated in Fig. \ref{flow_order}.
The dataset consists of three tasks that are essential components for automatic surgical perception and manipulation.
Besides, a three-tier annotation process is carefully designed to support learning-based model development of three tasks with uni- and multi-modality of data.
Details of the task and dataset design are presented in Section \ref{task_formulation} and \ref{Data_annotation}.

\subsection{Dataset collection}
In this work, we collect 21 videos of laparoscopic hysterectomy from Prince of Wales Hospital, Hong Kong, and then conduct pre-processing and annotation to build the dataset.
These 21 cases were performed from October to December 2018, using the Olympus imaging platform.
The usage of these surgical video data is approved by The Joint CUHK-NTEC Clinical Research Ethics Committee and the data are protected without disclosure of any personal information.

The 21 videos are recorded at 25 fps with a standard resolution of 1920 × 1080 pixels. 
The duration of videos ranges from 27 to 112 minutes due to the varying difficulties of the surgeries.
It should be noticed that the laparoscope is inevitably taken out from the abdomen when changing instruments or cleaning contamination on the lens, and the video clips during these periods are invalid for visual-based analysis and removed from our dataset to ensure a compact connection.
After pre-processing, the average duration is 66 minutes and the total duration is 1388 minutes, reaching a large-scale dataset with high-quality.

\subsection{Task formulation}
\label{task_formulation}
Three tasks are designed in the proposed AutoLaparo dataset:

\textbf{Task 1: surgical workflow recognition}.
This task focuses on the workflow analysis of laparoscopic hysterectomy procedure, which is fundamental to the scene understanding in surgery that helps to recognize current surgical phase and provide high-level information to the other two tasks \cite{twinanda2016endonet}.

\textbf{Task 2: laparoscope motion prediction}.
In MIS, surgeons need to operate within a proper FoV of the laparoscope, which in conventional surgery is held by an assistant.
Recently, some kinds of surgical robots have been designed for laparoscope motion control through human-machine interaction such as foot pedals, eye tracking \cite{fujii2018gaze}, etc., thereby liberating the assistant from tedious work but distracting surgeons from the surgical manipulation.
Therefore, state-of-the-art studies explored learning laparoscope motion patterns through image and video feedback \cite{li2022learning}, where appropriate datasets need to be developed for this vision-based automatic FoV control approach.

\textbf{Task 3: instrument and key anatomy segmentation}.
The segmentation of instrument and anatomy structure plays an important role in the realization of surgical automation, and the results of detection and segmentation can also serve high-level tasks, such as instrument tracking, pose estimation \cite{allan20183}, etc.

\subsection{Dataset annotation and statistics}
\label{Data_annotation}
Three sub-datasets are designed for the aforementioned three tasks and they are also highly-correlated by applying a three-tier annotation process.
The annotation is performed by a senior gynaecologist with more than thirty years of clinical experience and a specialist with three years of experience in hysterectomy.
The annotation results are proofread and cross-checked to ensure accuracy and consistency, and then stored in the dataset.

\subsubsection{Workflow annotation at video-level.}
Based on the clinical experience and domain knowledge \cite{blikkendaal2017surgical}, the 21 videos of laparoscopic hysterectomy are manually annotated into 7 phases: \textit{Preparation}, \textit{Dividing Ligament and Peritoneum}, \textit{Dividing Uterine Vessels and Ligament}, \textit{Transecting the Vagina}, \textit{Specimen Removal}, \textit{Suturing}, and \textit{Washing}. Fig. \ref{flow_order}(a) represents the order information and sample frames of each phase (P1-P7).
Specifically, since Phase 2 and 3 are performed symmetrically on the left and right sides of the uterus (denoted by A and B), the sequence of these two phases is related to the surgeon’s operating habits.
Some surgeons prefer to perform in a stage-by-stage sequence, while others prefer to handle the tissues on one side before proceeding to the other side.

\subsubsection{Motion annotation at clip-level.}
To further promote the automatic FoV control, we propose a sub-dataset containing selected clips that corresponds the laparoscope visual feedback to its motion mode.
In surgical videos, the manual laparoscope movements are usually momentary motions, lasting for $<$ 1 second, so that we denote the time $T$ as the beginning time of the motion. 
The main idea of this task is to predict the laparoscope motion at time $T$ using the visual information of the clip before that time, i.e., 0-$T$ (defined as “non-optimal view”).
For the clip after $T$ (defined as the “optimal view”), the video information will be provided as reference to present the scene after this motion.

Regarding the statistics, 300 clips are carefully selected from Phase 2-4 of the aforementioned 21 videos and each clip lasts for 10 seconds.
These clips contain typical laparoscope motions and the motion time $T$ is set as the fifth second as illustrated in Fig. \ref{flow_order}(b).
Seven types of motion modes are defined, including one static mode and six non-static mode: \textit{Up}, \textit{Down}, \textit{Left}, \textit{Right}, \textit{Zoom-in}, and \textit{Zoom-out}, and the number of video clips in each category are listed in Table \ref{laparoscope_motion_table}.

\begin{table}[t]
\begin{center}
\caption{Type and number of the clips with laparoscope motion annotation.}
\label{laparoscope_motion_table}
\setlength{\tabcolsep}{1.8mm}{
\begin{tabular}{c|ccccccc|c}
\hline
Motion Type & Up & Down & Left & Right & Zoom-in & Zoom-out & Static & All\\
\hline
Train set      & 14 & 25 & 18 & 12 & 30 & 17 & 54 & 170\\
Validation set & 3 & 4 & 10 & 3 & 15 & 12 & 10 & 57 \\
Test set       & 5 & 16 & 9 & 5 & 9 & 15 & 14 & 73\\
All Num.       & 22 & 45 & 37 & 20 & 54 & 44 & 78 & 300\\
\hline
\end{tabular}}
\end{center}
\end{table}

\begin{table*}[t]
\begin{center}
\caption{
Total number of presences in clips and frames, average number of pixels per frame, and total number of annotations of each class in the segmentation dataset.}
\label{seg_table}
\setlength{\tabcolsep}{1mm}{
\begin{tabular}{ll|ccc|cccc}
\hline
  \multicolumn{2}{c|}{ \multirow{2}*{Type}} & {Presence} & {Presence}  & {Average}  & \multicolumn{4}{c}{Annotation number} \\
\cline{6-9}
\multicolumn{2}{c|}{}&in Clips&in Frames&Pixels & All & Train set & Val. set & Test set \\
\hline
Anatomy & Uterus  & 164 &941&253968&		1016	& 663 &	214 & 139\\
Instrument&1-m	& 121 &575&57042&		577	& 371 &	107	& 99\\
&1-s    & 102 &439&92545&		442	& 291 &	89	& 62\\
&2-m	& 261 &1476&113569&		1481& 842 &	283	& 356\\
&2-s	& 228 &1071&137380&		1074& 630 &	206	& 238\\
&3-m	& 109 &518&59341&		538	& 287 &	92	& 159\\
&3-s	& 85 &365&94417&		373	& 189 &	81	& 103\\
&4-m	& 39 &200&7042&		202	& 104 &	50	& 48\\
&4-s	& 40 &233&179904&		233	& 124 &	55	& 54\\
\hline
All&	& 300 &1800&2073600& 5936 & 3501 &	1127 &	1258\\
\hline
\end{tabular}}
\end{center}
\end{table*}

\begin{figure} [t]
\centering 
\subfigure[Instrument-1]{
\includegraphics[width=0.19\linewidth]{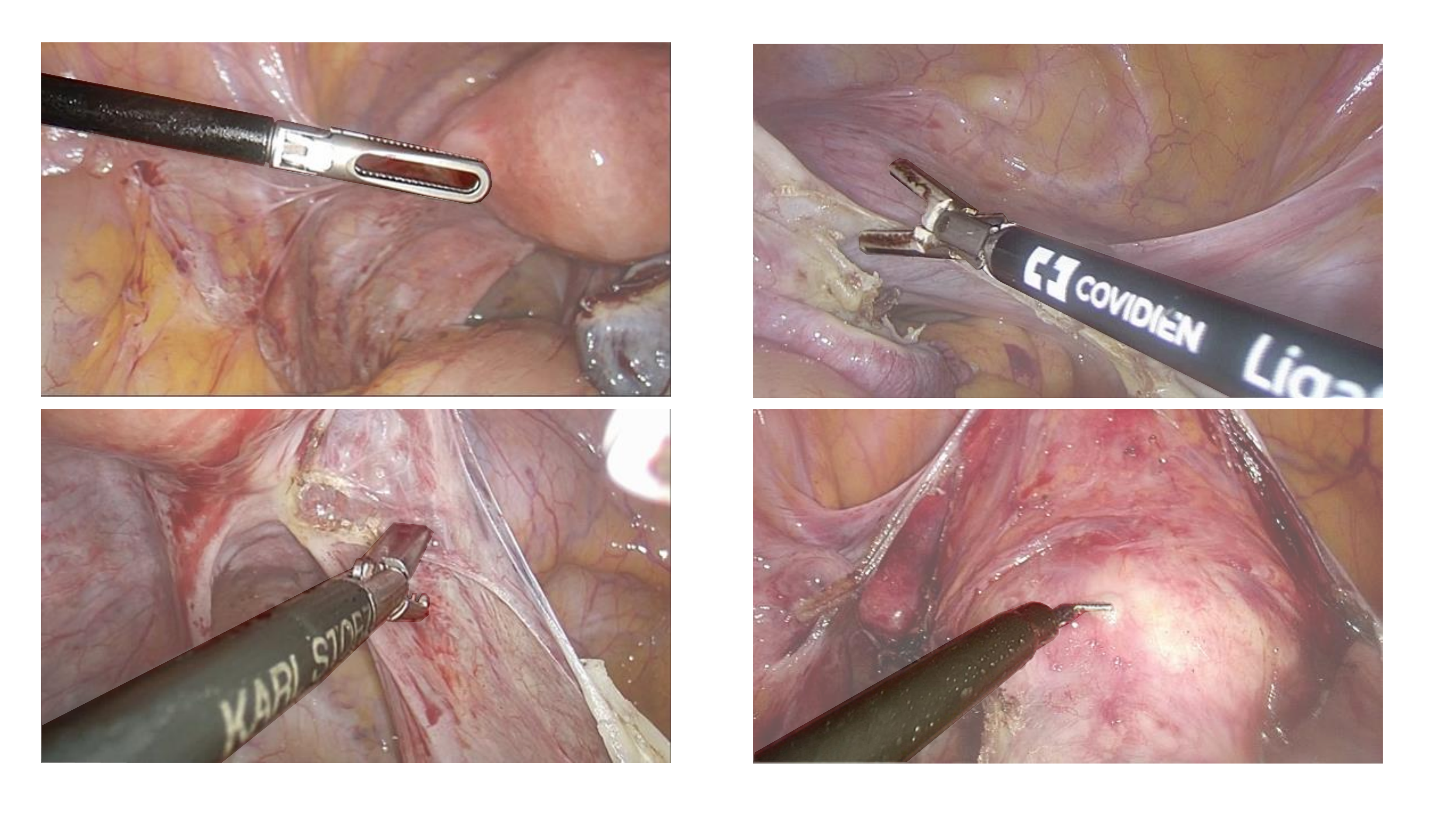}}
\hspace{-2mm}
%\vskip
\subfigure[Instrument-2]{
\includegraphics[width=0.19\linewidth]{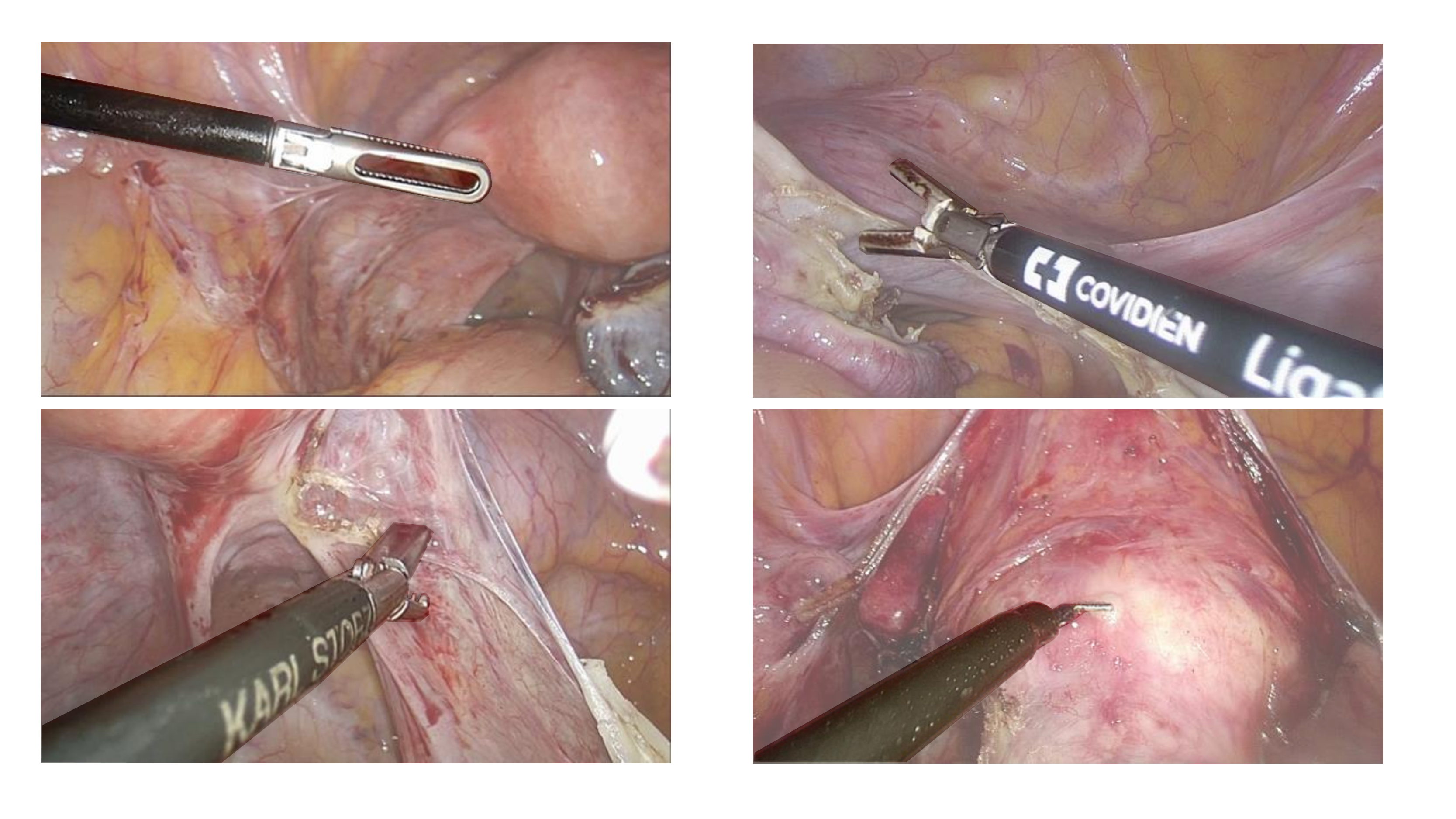}}
\hspace{-2mm}
%\vfill
\subfigure[Instrument-3]{
\includegraphics[width=0.19\linewidth]{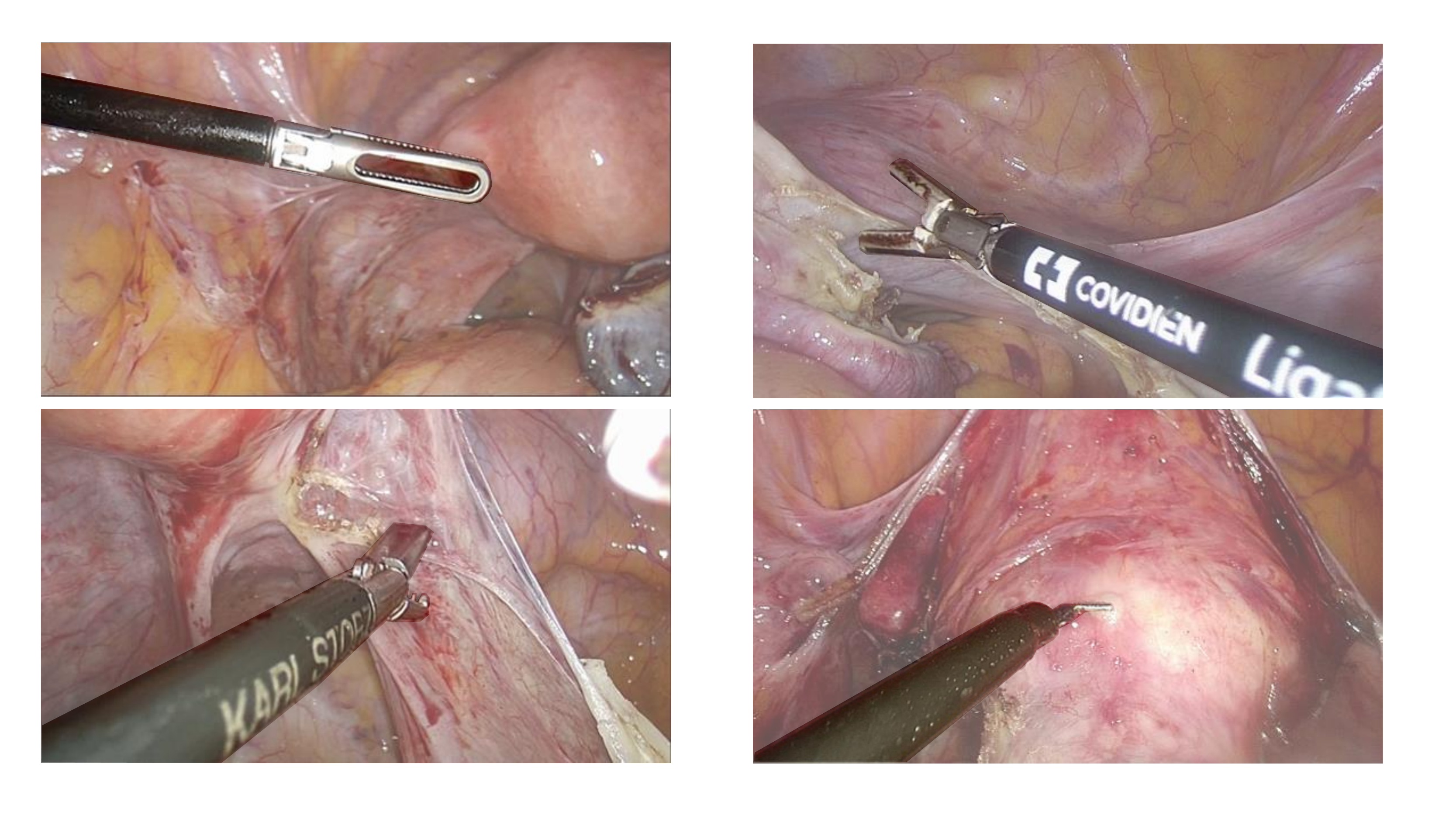}}
\hspace{-2mm}
\subfigure[Instrument-4]{
\includegraphics[width=0.19\linewidth]{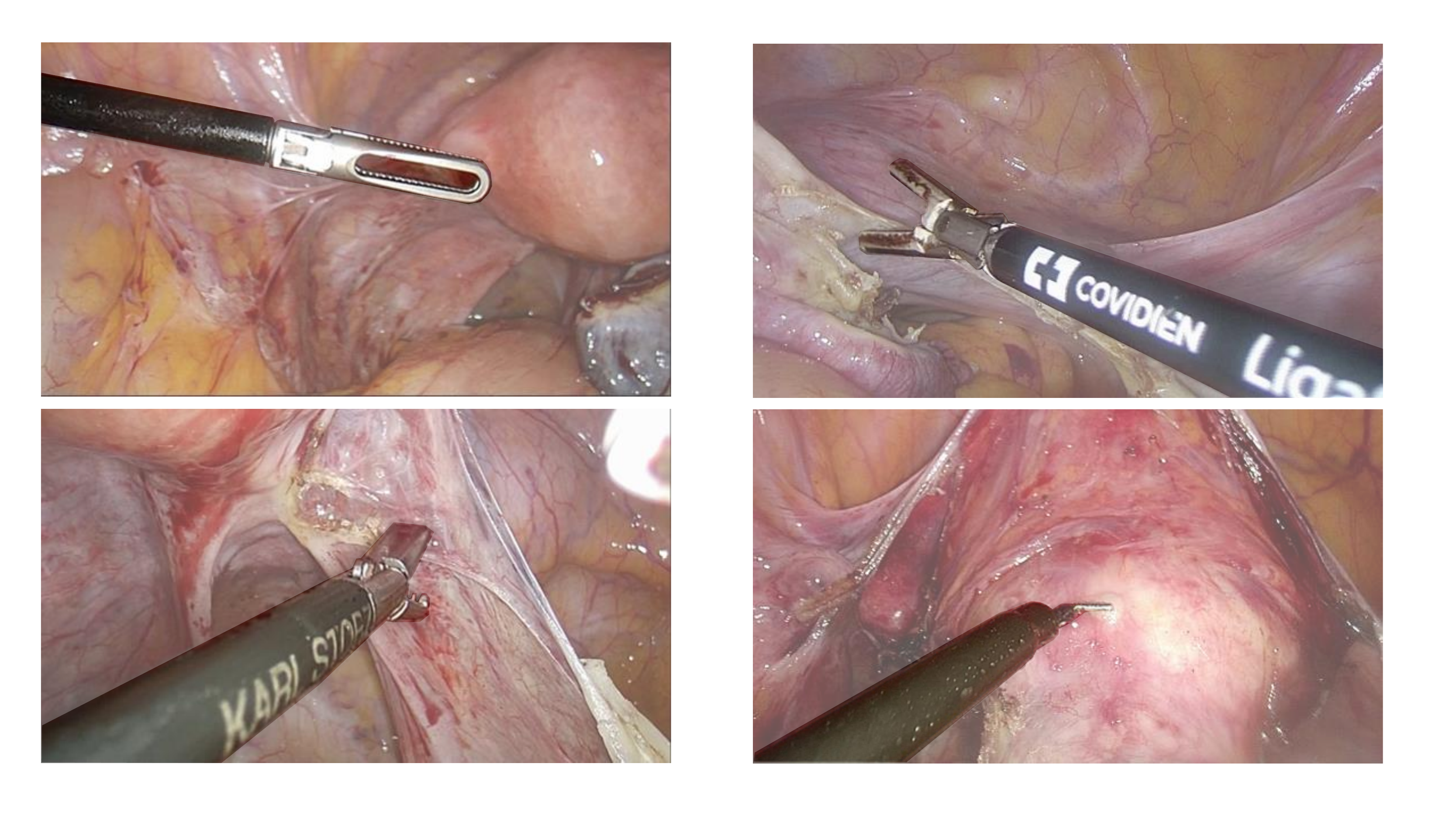}}
\hspace{-2mm}
%\vfill
\subfigure[Key anatomy]{
\includegraphics[width=0.19\linewidth]{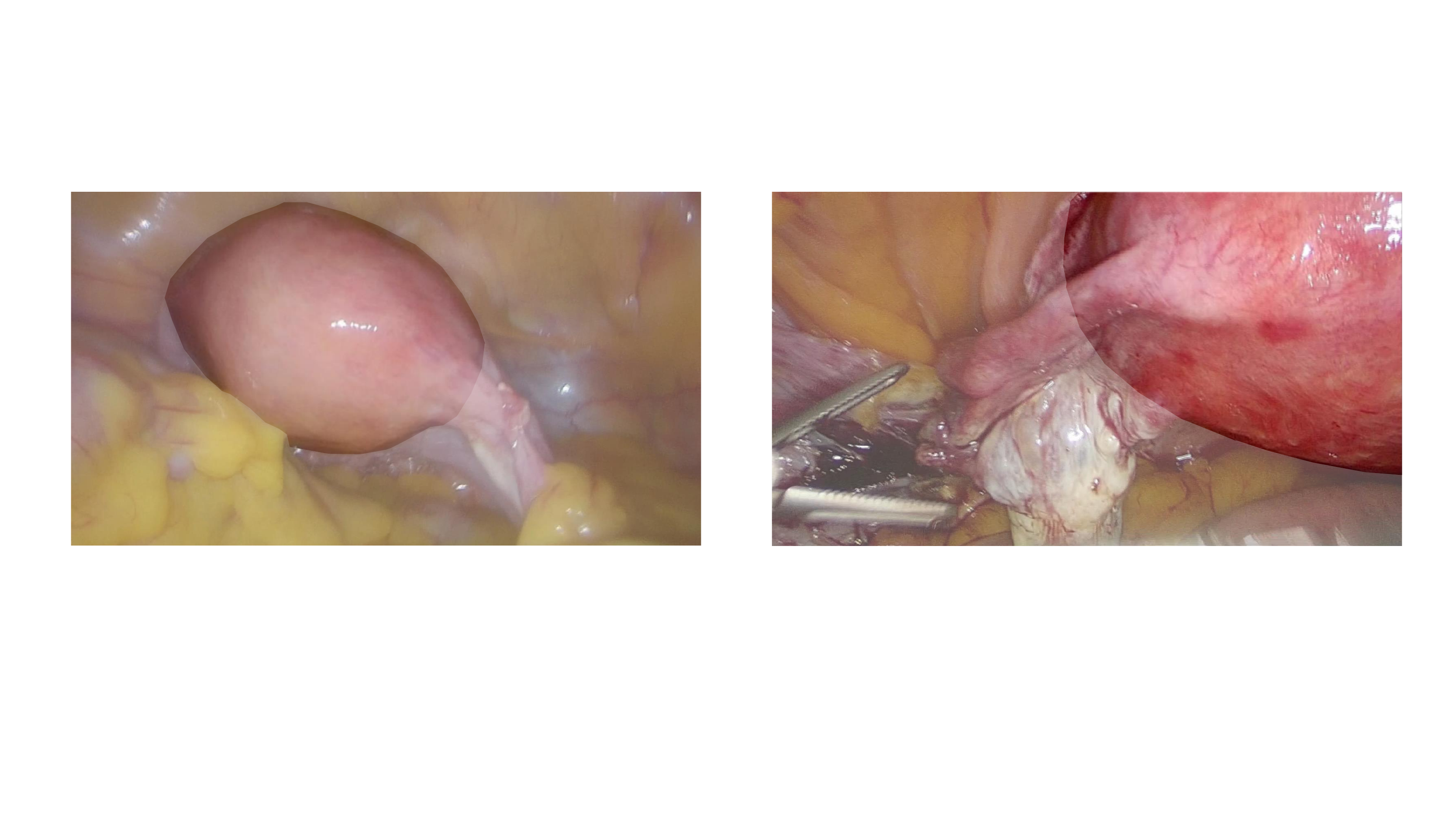}}
\caption{Illustrations of four instruments and the key anatomy (uterus) annotated in the AutoLaparo dataset. The background is blurred to highlight the object.}
\label{tools_fig}
\vspace{-1mm}
\end{figure}

\subsubsection{Segmentation annotation at frame-level.}
Based on the selected clips above, we further develop a sub-dataset with frame-level pixel-wise segmentation annotations of the surgical instrument and key anatomy. It provides ground truth for Task 3 and also supports Task 1 and 2 as an additional modality for advanced scene understanding.
Following the configuration of motion annotation, where the first five seconds of each clip are used as visual information to infer the motion at the fifth second, six frames are sampled at 1 fps, that are, frames at time $t$=0,1,2,3,4,5 (i.e., frame id 1,25,50,75,100,125), as illustrated in Fig. \ref{flow_order}.

There are four types of instruments and one key anatomy appearing in these frames: Instrument-1 \textit{Grasping forceps}, Instrument-2 \textit{LigaSure}, Instrument-3 \textit{Dissecting and grasping forceps}, Instrument-4 \textit{Electric hook}, and \textit{Uterus}, as shown in Fig. \ref{tools_fig}.
The annotations are performed with an open annotation tool called “LabelMe”~\cite{labelme2016} available online, following the settings of \textit{instrument part segmentation} \cite{allan20192017}, i.e., the shaft and manipulator of each instrument are annotated separately, as illustrated in Fig. \ref{flow_order}(c).
Finally, we reach a large-scale sub-dataset for segmentation containing 1800 annotated frames with 9 types and 5936 annotations.
Dataset split and statistics are presented in Table \ref{seg_table}, where ‘1-m’ denotes the manipulation of Instrument-1 and ‘1-s’ denotes its shaft.
It can also be observed that the Instrument-2, the main instrument in the surgery, appears most frequently, while the uterus appears in more than half of the frames.

\subsection{Dataset analysis and insights}
AutoLaparo is designed towards image-guided automation by achieving a comprehensive understanding of the entire surgical scene with three integrated tasks and data.
In specific, tool usage information is beneficial for recognizing phase and vice versa, so that these two kinds of annotation help to exploit the complementary information for Task 1 and 3.
Besides, the workflow and segmentation results provide rich information about the surgical scene that can help predict the laparoscope motion in Task 2 for further automatic FoV control.
In this way, the impact of each modality and the added value of combining several modalities for advanced perception can be explored with our integrated dataset.

\section{Experiments and Benchmarking Methods}
In this section, a set of experiments are performed for each task on AutoLaparo and benchmarks are provided for further model evaluation on this dataset.

\subsection{Workflow recognition}
\subsubsection{State-of-the-art methods.}

Four models are evaluated on workflow recognition task, that are, SV-RCNet \cite{jin2017sv}, TMRNet \cite{jin2021temporal}, TeCNO \cite{czempiel2020tecno}, and Trans-SVNet \cite{gao2021trans}.
SV-RCNet integrates visual and temporal dependencies in an end-to-end architecture.
TMRNet relates multi-scale temporal patterns with a long-range memory bank and a non-local bank operator.
TeCNO exploits temporal modelling with higher temporal resolution and large receptive field.
Trans-SVNet attempts to use Transformer to fuse spatial and temporal embeddings.

\subsubsection{Evaluation metrics and results.}
In AutoLaparo, 21 videos are divided into three subsets for training, validation, and test, containing 10, 4, and 7 videos, respectively.
The performances are comprehensively evaluated with four commonly-used metrics: accuracy (AC), precision (PR), recall (RE), and Jaccard (JA).
AC is defined as the percentage of correctly classified frames in the whole video, and PR, RE and JA are calculated by: ${\rm PR} = \frac{\mid{\rm GT}\cap{\rm P}\mid}{\mid{\rm P}\mid}$, ${\rm RE} = \frac{\mid{\rm GT}\cap{\rm P}\mid}{\mid{\rm GT}\mid}$, ${\rm JA} = \frac{\mid{\rm GT}\cap{\rm P}\mid}{\mid{\rm GT}\cup{\rm P}\mid}$, where GT and P denotes the ground truth and prediction set, respectively.

Experimental results are listed in Table \ref{flow_results_1}.
All approaches present satisfactory results with accuracies above 75\% on both validation and test sets.
The overall results present the potential to be applied to online automatic surgical workflow recognition and also used as guidance for the following tasks.

\begin{table}[t]
\begin{center}
\caption{Experimental results (\%) of four state-of-the-art models on Task 1.}
\label{flow_results_1}
\setlength{\tabcolsep}{2mm}{
\begin{tabular}{c|cccc|cccc}
\hline
\multirow{2}*{Method} &\multicolumn{4}{c|}{Validation Set}&\multicolumn{4}{c}{Test Set}\\
\cline{2-9}
{}& AC & PR & RE & JA &AC & PR & RE & JA \\
\hline
SV-RCNet & 77.22 & 72.28 & 65.73 & 51.64& 75.62 & 64.02 & 59.70 & 47.15\\
TMRNet & 80.96 & 79.11 & 67.36 & 54.95& 78.20 & 66.02 & 61.47 & 49.59 \\
TeCNO & 81.42 & 66.64 & 68.76 & 55.24& 77.27 & 66.92 & 64.60 & 50.67  \\
Trans-SVNet & 82.01 & 66.22 & 68.48 & 55.56 & 78.29 & 64.21 & 62.11 & 50.65 \\
\hline
\end{tabular}}
\end{center}
\end{table}

\subsection{Laparoscope motion prediction}
\subsubsection{State-of-the-art methods.}
In this task, we propose two methods to predict the laparoscope motion with uni-modality of visual input, i.e., the 5-second sub-clip before the motion time $T$.
The clips are downsampled from 25 fps to 3 fps to capture rich information from sequential frames, and in this way, 16 frames in each sub-clip (including the first and last frames) are input to the models.

In the first method, a ResNet-50 structure is developed to extract visual features of each frame, which are then averaged to represent each clip and fed into a fully-connected (FC) layer for motion mode prediction.
For the second method named ResNet-50+LSTM, a long short-term memory (LSTM) module is seamlessly integrated following the ResNet-50 backbone to exploit the long-range temporal information inherent in consecutive frames and output spatial-temporal embeddings.
Then the motion prediction results are generated from an FC layer connected to the LSTM module.

\subsubsection{Evaluation metrics and results.}
The metric used to assess the performance of the prediction results is the accuracy, which is defined as the percentage of the clips correctly classified into the ground truth in the entire dataset.

Experimental results are presented in Table \ref{motion_results_1}. The accuracies of these two models are 28.07\% and 29.82\% on validation set, and 26.03\% and 27.40\% on test set.
It can be seen that the second model achieves a higher accuracy value since it jointly learns both spatial representations and temporal motion patterns. 
The overall results need to be improved and advanced methods combining multi-modality data could be proposed for this important but challenging task.

\begin{table}[t]
\begin{center}
\caption{Experimental results (\%) of two proposed methods on Task 2.}
\label{motion_results_1}
\setlength{\tabcolsep}{6mm}{
\begin{tabular}{c|ccc}
\hline
Method  & Validation set & Test set  \\
\hline
Resnet-50  & 28.07 &  26.03\\
Resnet-50+LSTM  & 29.82 & 27.40\\
\hline
\end{tabular}}
\end{center}
\end{table}

\begin{table}[t]
\begin{center}
\caption{Experimental results (\%) of three state-of-the-art models on Task 3.}
\label{segmentation_results_1}
\setlength{\tabcolsep}{4mm}{
\begin{tabular}{c|ccc}
\hline
Method & Backbone & box AP & mask AP \\%& FPS\\
\hline
Mask R-CNN & ResNet-50 & 52.10 & 51.70  \\%& 7.8\\
YOLACT & ResNet-50 & 48.16 & 54.09 \\%& 25.6\\
YolactEdge &	ResNet-50 & 47.04 & 52.58 \\%& 30.2\\
\hline
\end{tabular}}
\end{center}
\end{table}

\subsection{Instrument and key anatomy segmentation}

\subsubsection{State-of-the-art methods.}

Three models for object detection and instance segmentation are evaluated to give benchmarks: Mask R-CNN~\cite{he2017mask}, a representative two-stage approach known as a competitive baseline model; YOLACT~\cite{bolya2019yolact}, the first algorithm that realizes real-time instance segmentation, reaching a considerable speed of 30 fps with satisfactory results; and YolactEdge~\cite{liu2020yolactedge} that yields a speed-up over existing real-time methods and also competitive results.

\subsubsection{Evaluation metrics and results.}
The experimental results are reported using the standard COCO metrics where the average precision (AP) is calculated by averaging the intersection over union (IoU) with 10 thresholds of 0.5 : 0.05 : 0.95 for both bounding boxes (box AP) and segmentation masks (mask AP).

For fair comparison of baseline models, We conduct experiments with a same backbone ResNet-50 and the results on the test set are presented in Table \ref{segmentation_results_1}.
The box AP are 52.10\%, 48.16\%, and 47.04\% for the three methods respectively.
For segmentation, Mask R-CNN gives mask AP of 51.70\%, YOLACT of 54.09\%, and YolactEdge of 52.58\%, showing promising potential for practical applications.

\section{Conclusion and Future Work}

In this paper, we propose AutoLaparo, the first integrated dataset to facilitate visual perception and image-guided automation in laparoscopic hysterectomy.
The dataset consists of raw videos and annotations towards multi-task learning and three tasks are defined: surgical workflow recognition, laparoscope motion prediction, and instrument and key anatomy segmentation, which are not independent but also highly correlated for advanced scene understanding. 
Besides, experiments are performed on each task and benchmarks are presented as reference for the future model development and evaluation on this dataset.

The remaining challenges for this dataset are how to develop models to improve the performances of each task with uni-modality data and achieve multi-task learning by combining various modalities.
By making this dataset publicly available, we and the whole community will together dedicate to addressing these issues in the future based on our integrated data and advanced learning strategies.
In addition, we plan to extend this dataset by expanding the amount of images and videos and defining more tasks towards vision-based applications.

\subsubsection{Acknowledgement.}
This work is supported in part by Shenzhen Portion of Shenzhen-Hong Kong Science and Technology Innovation Cooperation Zone under HZQB-KCZYB-20200089,  in part of the HK RGC under T42-409/18-R and 14202918,  in part by the Multi-Scale Medical Robotics Centre, InnoHK, and in part by the VC Fund 4930745 of the CUHK T Stone Robotics Institute.

% \clearpage
\bibliographystyle{splncs04}
% \normalem
\bibliography{reference}

\begin{thebibliography}{10}
\providecommand{\url}[1]{\texttt{#1}}
\providecommand{\urlprefix}{URL }
\providecommand{\doi}[1]{https://doi.org/#1}

\bibitem{allan20202018}
Allan, M., Kondo, S., Bodenstedt, S., Leger, S., Kadkhodamohammadi, R., Luengo,
  I., Fuentes, F., Flouty, E., Mohammed, A., Pedersen, M., et~al.: 2018 robotic
  scene segmentation challenge. arXiv preprint arXiv:2001.11190  (2020)

\bibitem{allan20183}
Allan, M., Ourselin, S., Hawkes, D.J., Kelly, J.D., Stoyanov, D.: 3-d pose
  estimation of articulated instruments in robotic minimally invasive surgery.
  IEEE transactions on medical imaging  \textbf{37}(5),  1204--1213 (2018)

\bibitem{allan20192017}
Allan, M., Shvets, A., Kurmann, T., Zhang, Z., Duggal, R., Su, Y.H., Rieke, N.,
  Laina, I., Kalavakonda, N., Bodenstedt, S., et~al.: 2017 robotic instrument
  segmentation challenge. arXiv preprint arXiv:1902.06426  (2019)

\bibitem{barbash2010new}
Barbash, G.I.: New technology and health care costs--the case of robot-assisted
  surgery. The New England journal of medicine  \textbf{363}(8), ~701 (2010)

\bibitem{bihlmaier2014automated}
Bihlmaier, A., Woern, H.: Automated endoscopic camera guidance: A
  knowledge-based system towards robot assisted surgery. In: ISR/Robotik 2014;
  41st International Symposium on Robotics. pp.~1--6. VDE (2014)

\bibitem{blikkendaal2017surgical}
Blikkendaal, M.D., Driessen, S.R., et~al.: Surgical flow disturbances in
  dedicated minimally invasive surgery suites: an observational study to assess
  its supposed superiority over conventional suites. Surgical endoscopy
  \textbf{31}(1),  288--298 (2017)

\bibitem{bolya2019yolact}
Bolya, D., Zhou, C., Xiao, F., Lee, Y.J.: Yolact: Real-time instance
  segmentation. In: Proceedings of the IEEE/CVF International Conference on
  Computer Vision. pp. 9157--9166 (2019)

\bibitem{czempiel2020tecno}
Czempiel, T., Paschali, M., Keicher, M., Simson, W., Feussner, H., Kim, S.T.,
  Navab, N.: Tecno: Surgical phase recognition with multi-stage temporal
  convolutional networks. In: International Conference on Medical Image
  Computing and Computer-Assisted Intervention. pp. 343--352. Springer (2020)

\bibitem{dergachyova2016automatic}
Dergachyova, O., Bouget, D., Huaulm{\'e}, A., Morandi, X., Jannin, P.:
  Automatic data-driven real-time segmentation and recognition of surgical
  workflow. International journal of computer assisted radiology and surgery
  \textbf{11}(6),  1081--1089 (2016)

\bibitem{farquhar2002hysterectomy}
Farquhar, C.M., Steiner, C.A.: Hysterectomy rates in the united states
  1990--1997. Obstetrics \& gynecology  \textbf{99}(2),  229--234 (2002)

\bibitem{fujii2018gaze}
Fujii, K., Gras, G., Salerno, A., Yang, G.Z.: Gaze gesture based human robot
  interaction for laparoscopic surgery. Medical image analysis  \textbf{44},
  196--214 (2018)

\bibitem{gao2021trans}
Gao, X., Jin, Y., Long, Y., Dou, Q., Heng, P.A.: Trans-svnet: accurate phase
  recognition from surgical videos via hybrid embedding aggregation
  transformer. In: International Conference on Medical Image Computing and
  Computer-Assisted Intervention. pp. 593--603. Springer (2021)

\bibitem{gao2014jhu}
Gao, Y., Vedula, S.S., Reiley, C.E., Ahmidi, N., Varadarajan, B., Lin, H.C.,
  Tao, L., Zappella, L., B{\'e}jar, B., Yuh, D.D., et~al.: Jhu-isi gesture and
  skill assessment working set (jigsaws): A surgical activity dataset for human
  motion modeling. In: MICCAI workshop: M2cai. vol.~3, p.~3 (2014)

\bibitem{grammatikopoulou2021cadis}
Grammatikopoulou, M., Flouty, E., Kadkhodamohammadi, A., Quellec, G., Chow, A.,
  Nehme, J., Luengo, I., Stoyanov, D.: Cadis: Cataract dataset for surgical
  rgb-image segmentation. Medical Image Analysis  \textbf{71},  102053 (2021)

\bibitem{he2017mask}
He, K., Gkioxari, G., Doll{\'a}r, P., Girshick, R.: Mask r-cnn. In: Proceedings
  of the IEEE international conference on computer vision. pp. 2961--2969
  (2017)

\bibitem{huaulme2022peg}
Huaulm{\'e}, A., Harada, K., Nguyen, Q.M., Park, B., Hong, S., Choi, M.K.,
  Peven, M., Li, Y., Long, Y., Dou, Q., et~al.: Peg transfer workflow
  recognition challenge report: Does multi-modal data improve recognition?
  arXiv preprint arXiv:2202.05821  (2022)

\bibitem{jin2017sv}
Jin, Y., Dou, Q., Chen, H., Yu, L., Qin, J., Fu, C.W., Heng, P.A.: Sv-rcnet:
  workflow recognition from surgical videos using recurrent convolutional
  network. IEEE transactions on medical imaging  \textbf{37}(5),  1114--1126
  (2017)

\bibitem{jin2021temporal}
Jin, Y., Long, Y., Chen, C., Zhao, Z., Dou, Q., Heng, P.A.: Temporal memory
  relation network for workflow recognition from surgical video. IEEE
  Transactions on Medical Imaging  (2021)

\bibitem{leibetseder2020glenda}
Leibetseder, A., Kletz, S., Schoeffmann, K., Keckstein, S., Keckstein, J.:
  Glenda: Gynecologic laparoscopy endometriosis dataset. In: International
  Conference on Multimedia Modeling. pp. 439--450. Springer (2020)

\bibitem{leibetseder2018lapgyn4}
Leibetseder, A., Petscharnig, S., Primus, M.J., Kletz, S., M{\"u}nzer, B.,
  Schoeffmann, K., Keckstein, J.: Lapgyn4: a dataset for 4 automatic content
  analysis problems in the domain of laparoscopic gynecology. In: Proceedings
  of the 9th ACM Multimedia Systems Conference. pp. 357--362 (2018)

\bibitem{li2022learning}
Li, B., Lu, B., Wang, Z., Zhong, B., Dou, Q., Liu, Y.: Learning laparoscope
  actions via video features for proactive robotic field-of-view control. IEEE
  Robotics and Automation Letters  (2022)

\bibitem{liu2020yolactedge}
Liu, H., Soto, R.A.R., Xiao, F., Lee, Y.J.: Yolactedge: Real-time instance
  segmentation on the edge. arXiv preprint arXiv:2012.12259  (2020)

\bibitem{maier2022surgical}
Maier-Hein, L., Eisenmann, M., Sarikaya, D., M{\"a}rz, K., Collins, T.,
  Malpani, A., Fallert, J., Feussner, H., Giannarou, S., Mascagni, P., et~al.:
  Surgical data science--from concepts toward clinical translation. Medical
  image analysis  \textbf{76},  102306 (2022)

\bibitem{maier2017surgical}
Maier-Hein, L., Vedula, S.S., Speidel, S., Navab, N., Kikinis, R., Park, A.,
  Eisenmann, M., Feussner, H., Forestier, G., et~al.: Surgical data science for
  next-generation interventions. Nature Biomedical Engineering  \textbf{1}(9),
  691--696 (2017)

\bibitem{merrill2008hysterectomy}
Merrill, R.M.: Hysterectomy surveillance in the united states, 1997 through
  2005. Medical Science Monitor  \textbf{14}(1),  CR24--CR31 (2008)

\bibitem{nakawala2019deep}
Nakawala, H., Bianchi, R., Pescatori, L.E., De~Cobelli, O., Ferrigno, G.,
  De~Momi, E.: “deep-onto” network for surgical workflow and context
  recognition. International journal of computer assisted radiology and surgery
   \textbf{14}(4),  685--696 (2019)

\bibitem{sarikaya2017detection}
Sarikaya, D., Corso, J.J., Guru, K.A.: Detection and localization of robotic
  tools in robot-assisted surgery videos using deep neural networks for region
  proposal and detection. IEEE transactions on medical imaging  \textbf{36}(7),
   1542--1549 (2017)

\bibitem{taylor2008medical}
Taylor, R.H., Kazanzides, P.: Medical robotics and computer-integrated
  interventional medicine. In: Biomedical Information Technology, pp. 393--416.
  Elsevier (2008)

\bibitem{topol2019high}
Topol, E.J.: High-performance medicine: the convergence of human and artificial
  intelligence. Nature medicine  \textbf{25}(1),  44--56 (2019)

\bibitem{tsui2013minimally}
Tsui, C., Klein, R., Garabrant, M.: Minimally invasive surgery: national trends
  in adoption and future directions for hospital strategy. Surgical endoscopy
  \textbf{27}(7),  2253--2257 (2013)

\bibitem{twinanda2016endonet}
Twinanda, A.P., Shehata, S., Mutter, D., Marescaux, J., De~Mathelin, M., Padoy,
  N.: Endonet: a deep architecture for recognition tasks on laparoscopic
  videos. IEEE transactions on medical imaging  \textbf{36}(1),  86--97 (2016)

\bibitem{labelme2016}
Wada, K.: {labelme: Image Polygonal Annotation with Python}.
  \url{https://github.com/wkentaro/labelme} (2016)

\bibitem{zadeh2020surgai}
Zadeh, S.M., Francois, T., Calvet, L., Chauvet, P., Canis, M., Bartoli, A.,
  Bourdel, N.: Surgai: deep learning for computerized laparoscopic image
  understanding in gynaecology. Surgical endoscopy  \textbf{34}(12),
  5377--5383 (2020)

\end{thebibliography}

\end{document}